\documentclass[runningheads]{llncs}
\usepackage{graphicx}
\usepackage{epsfig}
\usepackage{amsfonts}

\usepackage{hyperref}
\hypersetup{colorlinks=true, allcolors=blue, urlcolor=purple}
\usepackage{soul}%
\usepackage[colorinlistoftodos]{todonotes}
\usepackage{xargs}%
\usepackage{float}
\usepackage{wrapfig}
\usepackage{multirow}
\usepackage[symbol]{footmisc}

\begin{document}
\title{Interactive Learning for Semantic Segmentation in Earth Observation }
%
%
\author{Gaston Lenczner\inst{1,2, *}\and
Adrien Chan-Hon-Tong\inst{1}\and
Nicola Luminari\inst{2}\and \\Bertrand Le Saux\inst{3}\and Guy Le Besnerais\inst{1}}
\authorrunning{G. Lenczner et al.}
%
\institute{ONERA/DTIS, Universit{é} Paris-Saclay, F-91123 Palaiseau, France\\ \email{name.surname@onera.fr}\and Delair, FR-31400 Toulouse, France\\ \email{name.surname@delair.aero}\and ESA/ESRIN, $\Phi$-Lab, I-00044 Frascati (RM), Italy\\\email{bls@ieee.com}}
\maketitle              
\begin{abstract}
Dense pixel-wise classification maps output by deep neural networks are of extreme importance for scene understanding. However, these maps are often partially inaccurate due to a variety of possible factors. Therefore, we propose to interactively refine them within a framework named DISCA (Deep Image Segmentation with Continual Adaptation). It consists of continually adapting a neural network to a target image using an interactive learning process with sparse user annotations as ground-truth. We show through experiments on three datasets using synthesized annotations the benefits of the approach, reaching an IoU improvement up to 4.7\% for ten sampled clicks. Finally, we exhibit that our approach can be particularly rewarding when it is faced to additional issues such as domain adaptation.

\end{abstract}

\section{Introduction}

\footnotetext[1]{Corresponding author \href{mailto:gaston.lenczner@delair.aero}{gaston.lenczner@delair.aero}}

As massive amounts of data are produced everyday coming from multiple sources such as drones or satellites~\cite{torres2012gmes}, Earth Observation (EO) data plays a central role in the way we understand our planet. Semantic segmentation, the task of classifying an image pixel-wise, is of primary importance in EO for various purposes such as land-cover classification and is now efficiently addressed by deep neural networks. However, there is no theoretical guarantee of the performances of these networks and they indeed may fail in practice. 
Besides, they often get worse when they are faced to additional constraints such as limited training database~\cite{milan2018semantic}, flawed labels~\cite{heller2018imperfect} or domain shifts~\cite{maggiori2017dataset} (different weather conditions, different geographical positions, different seasons, ...). In many situations, datasets deal with several, if not all, of these issues. Therefore, mistake-free neural networks are still an utopia in such scenarios. 

Depending on the applications, these errors can be controlled or tolerable and data can thus be processed fully automatically at large scale. However, they can also be unacceptable and data processing then needs to be controlled by a human operator who can certify the results in a semi-automatic way~\cite{le_saux_iros2013}. In practice, the user can work hand-in-hand with machine learning models which have been previously trained on large amounts of data.
Indeed, as shown by DIOS~\cite{xu2016deep} or DISIR~\cite{lenczner_disir_2020}
, a user can guide a neural network to perform segmentation tasks with clicked annotations given as inputs to the algorithm. Since these approaches do not modify the weights of the networks, their guidance is spatially localized around the annotations.
Therefore, we propose in the present paper Deep Image Segmentation with Continual Adaptation~(DISCA), a semantic segmentation framework to interactively retrain a neural network to enhance its performances on EO images at an image level using the annotations as a sparse reference.
Interestingly, remote sensing seems to be a more relevant setting than natural images for this kind of interactive learning. Indeed, there can be a large variability within a  natural image which makes it hard to exploit information provided by an annotation. Inversely, within a remote sensing image, variability tends to be low while the image tends to be large, allowing to take full advantage of the information provided by the user.

%
Our present contribution is the following. We introduce DISCA, assert its efficiency on three remote sensing datasets and also show that it can be particularly relevant in a domain shift context.


\section{State of the art}


\paragraph{Weakly supervised learning} aims at making algorithms learn on data with flawed or partial labels. In semantic segmentation, these labels can then take many forms such as single points~\cite{bearman2016s,chan2018object}, bounding-boxes~\cite{dai2015boxsup} or scribbles~\cite{lin2016scribblesup}. Weak supervision has begun to receive a growing attention in the remote sensing community as labels can be costly to acquire at large scale. ~\cite{caye2019guided} proposed an iterative training process based on weak supervision for the task of change detection. Our work is closely related to weak supervised learning as the user annotation maps can be seen as sparse ground-truth maps. Differently than previous works, we only fine-tune  in a weakly-supervised fashion the networks otherwise standardly trained. 

\paragraph{Continual learning} defines the ability of a learning algorithm to continuously learn from a stream of data~\cite{parisi2019continual}. It has been approached in remote sensing for semantic segmentation at large scale with new labels appearing over time~\cite{tasar2019incremental}. 
Continual learning often implies learning through multiple tasks. This applies to our work if we consider standard semantic segmentation as the first task and then learning from the target image and the annotations as the second one. We also face a challenge inherent to reinforcement learning~\cite{kaelbling1996reinforcement} where an agent (our network) interacts with a dynamic environment (the user annotations). Indeed, we must pay particular attention to the risk of classifier degradation.

\paragraph{Interactive segmentation} intends to interactively segment an image into foreground and background pixels with user annotations. It was initially addressed using graph-cut based methods~\cite{rother2004grabcut} and now mostly by deep neural networks which take as inputs a concatenation of the RGB image and user annotations~\cite{xu2016deep}. In~\cite{kontogianni2019continuous}, the authors use the annotations as sparse ground truth maps to interactively adapt the neural network to a specific object. Multi-class interactive segmentation broadens interactive segmentation to correct multi-class segmentation maps. ~\cite{agustsson2019interactive} proposed a neural network which takes as input a concatenation of the image and the extreme points of each instance in the scene and then lets a user correct the proposed multi-class segmentation using scribbles. We do not assume such extreme point map availability as it is costly to acquire in a remote sensing image with many potential objects. In our previous work~\cite{lenczner_disir_2020}, we extended~\cite{xu2016deep} to interactively refine remote sensing segmentation maps. We now draw inspiration from~\cite{kontogianni2019continuous} to the same purpose. 

\section{Methodology}
\begin{figure}[h]
\begin{minipage}[t]{1\linewidth}
\centering{\epsfig{figure=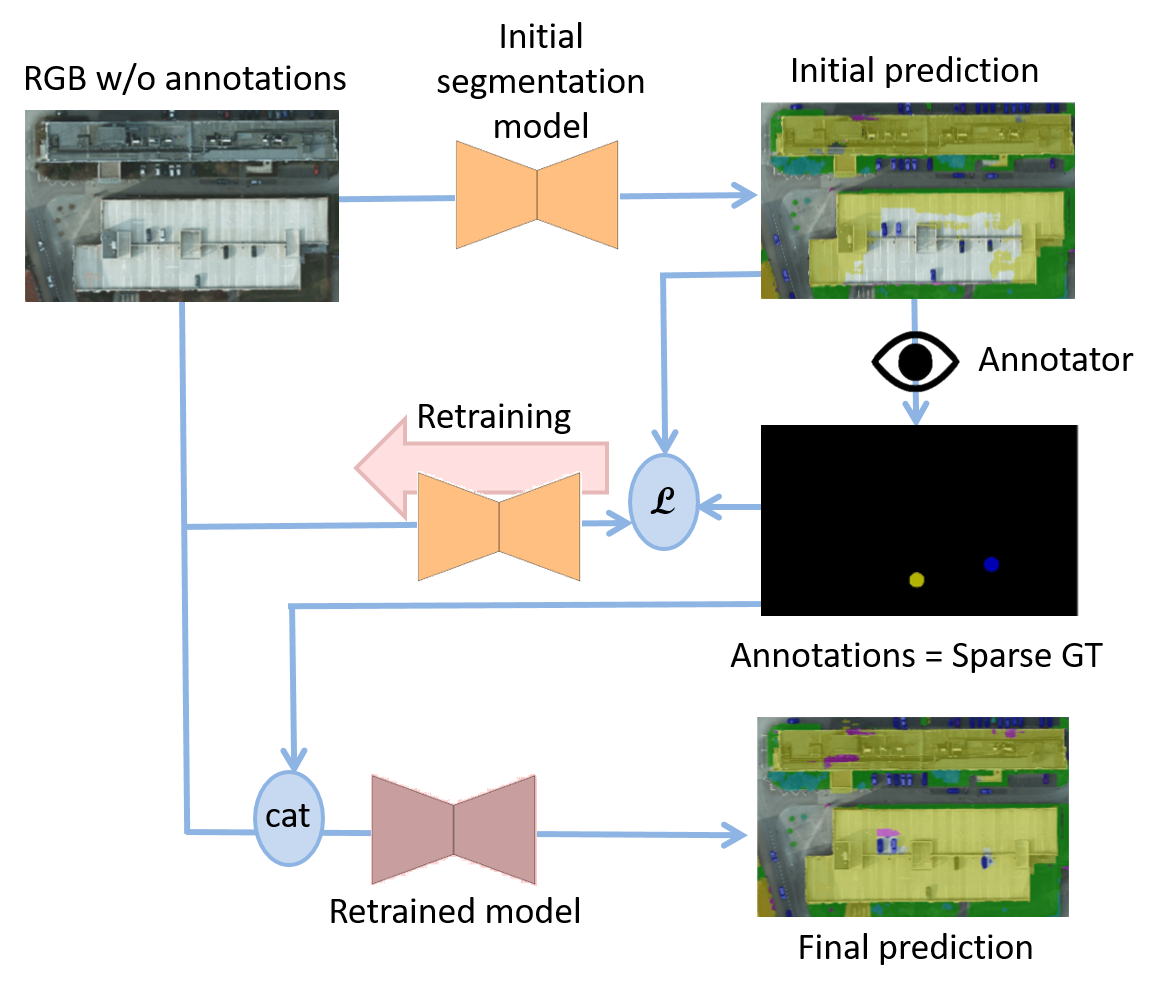,width=.85\linewidth}}
\end{minipage}
	\caption{Visual abstract of DISCA method. Up: initial prediction, middle: retraining using annotations as a sparse ground truth, bottom: final prediction using the retrained model in a DISIR~\cite{lenczner_disir_2020} mode.}
\label{fig:overview}
\end{figure}

We rely on DISIR\footnote{\href{https://github.com/delair-ai/DISIR}{https://github.com/delair-ai/DISIR}}~\cite{lenczner_disir_2020} since it is, to the best of our knowledge, the only existing open-source work addressing multi-class interactive segmentation using deep learning in remote sensing. Hence, our network takes  as input a concatenation of the RGB image and of the user annotations. Initially under the form of clicked points, these annotations are encoded using distance transforms into a $N$-dimensional tensor where $N$ is the cardinal of the label space. Sampled randomly from the ground-truth during an initial classic training phase, they are then used by the neural network as guidance to enhance its initial predictions. Since only the network's inputs are modified and not its parameters, the information provided by the annotations does not improve the predictions in the entire image. 
We propose to bypass this locality constraint by retraining the network with a few back-propagation cycles per annotation. 


To fully benefit from the annotations at the image scale, as shown in Figure~\ref{fig:overview}, we use them as a sparse ground-truth to interactively retrain the network using a cross entropy loss on these annotated pixels. We note \textbf{f} to represent the neural network parameterized by $\mathbf{\theta}$ and \textbf{x} its inputs. As only a few pixels are annotated among the millions ones that usually compose a remote sensing image, the ground-truth maps are extremely sparse. In order to deal with this problem and avoid over-fitting, we follow~\cite{kontogianni2019continuous} by using the initial prediction $\mathbf{p_0} = f(\mathbf{x}, \theta_0)$ for regularization. This regularization consists of adding a term of cross-entropy loss using the original prediction as ground-truth in order to prevent the model from making a prediction too different from the initial one. 
Differently from the source paper, we use a regularization function based on a  $L^1$ loss instead of a more permissive cross entropy loss and we do not add an additional regularization over the network parameters. Therefore, our loss during the interactive learning process is defined as follows:
\begin{equation}
\label{eq:loss}
\mathcal{L}(\mathbf{x}, \mathbf{c}, \mathbf{p_0}; \mathbf{\theta}) = \frac{\mathbf{1}_{[\mathbf{c=-1}]}}{\|\mathbf{1}_{[\mathbf{c=-1}]}\|_1}\left\{-\sum\limits_{i=1}^{N}\mathbf{c}_i\log\left(\mathbf{f}_i(\mathbf{x};\mathbf{\theta})\right)\right\}+\|\mathbf{f}(\mathbf{x}; \mathbf{\theta})-\mathbf{p_0}\|_1
\end{equation}
where \textbf{1} represents the indicator function and \textbf{c} the sparse annotated pixels. In details, \textbf{c} takes its values in \{-1, 0, 1\}. For the pixels annotated as belonging to class $i$, $\mathbf{c}_i=1$ and $\mathbf{c}_j=0$ for all $j\neq i$. For the unannotated pixels, $\mathbf{c}_i=-1$ for all $i$ in $\{1,\dots,N\}$. 

The DISIR mechanism is only used in the last inference. Indeed, the fine-tuning DISCA mechanism works on the image only. In other words, during the interactive training process, the annotations are not concatenated with the RGB image at the input of the network. This prevents the network from over-fitting on them. 
Fortunately, this does not make the network forget how to use the annotations for guidance.

\section{Experiments}
\subsection{Experimental setup}
In this section, we study through experiments the scope of our approach and compare it to~\cite{lenczner_disir_2020}. We experiment on three semantic segmentation remote sensing datasets: the INRIA Aerial Image Labelling dataset~\cite{maggiori2017dataset} composed of two classes (\textit{buildings} and \textit{not buildings}) and covering more than 800 $\textrm{km}^2$ in  different cities at a 30cm resolution, the Aerial Imagery for Roof Segmentation (AIRS) dataset~\cite{chen2018aerial} composed of the same two classes and covering  457 $\textrm{km}^2$ in New-Zealand at a 7.5cm resolution and the ISPRS Potsdam dataset~\cite{rottensteiner2012isprs} composed of 6 classes (\textit{impervious surface, buildings, low vegetation, tree, car} and \textit{clutter}) covering 3~$\textrm{km}^2$ on Potsdam at a 5cm resolution. The initial training sets are divided into a smaller training set and a validation set with a ratio 80\%-20\%. This allows to synthesise annotations to evaluate the framework. 
The images are tiled into patches of size $512\times 512$ with an overlap of size 128 to be processed.

We use a neural network based on the LinkNet~\cite{chaurasia2017linknet} architecture - but the approach is independent from the neural network backbone. To evaluate the refinement performances on the validation sets, we sample 10 annotations from the ground-truth maps in the largest wrongly predicted areas, adapt the networks in an image-wise fashion and measure the Intersection over Union (IoU) evolution. During the interactive learning phase, we optimize the weights using 10 stochastic gradient descent passes with a learning rate of $2e^{-7}$ and minimize the loss defined in Eq.~\ref{eq:loss}.

\subsection{Single datasets experiments}

\begin{table}[t]
    \caption{Mean IoU obtained before and after the interactive processes}
    \label{tab:results}
    \centering
    \begin{minipage}[t]{0.3\linewidth}
        \begin{tabular}{c|c|c}
          & DISIR & DISCA\\\hline
        Before & \multicolumn{2}{c}{70.6} \\ \hline
        After & 71.3 & \textbf{72.2}
    \end{tabular}
    \vspace{0.2cm}
    
    \centering\small{(a) ISPRS}
    \end{minipage}\hfill
    \begin{minipage}[t]{0.3\linewidth}
        \begin{tabular}{c|c|c}
          & DISIR & DISCA\\\hline
        Before &\multicolumn{2}{c}{85.4}\\ \hline
        After & 86.4 & \textbf{86.5}
    \end{tabular}
    \vspace{0.2cm}
    
    \centering\small{(b) INRIA}
    \end{minipage}\hfill
    \begin{minipage}[t]{0.3\linewidth}
        \begin{tabular}{c|c|c}
          & DISIR & DISCA\\\hline
        Before & \multicolumn{2}{c}{85.9}  \\ \hline
        After & 89.5 & \textbf{90.6}
    \end{tabular}
    \vspace{0.2cm}
    
    \centering\small{(c) AIRS}
     \end{minipage}
    \end{table}
    
\newcommand\x{0.24}
\begin{figure}[h]
\begin{minipage}[t]{\x\linewidth}
\centering{\epsfig{figure=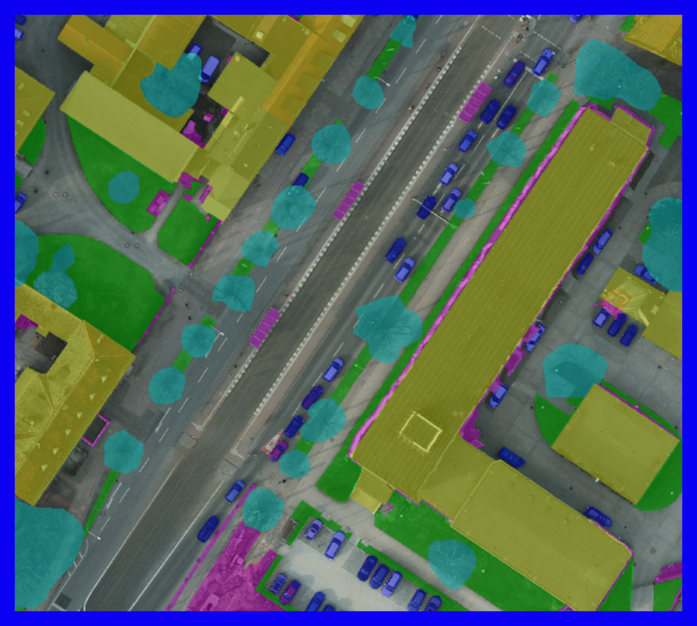,width=\linewidth}}
\end{minipage}\hfill
\begin{minipage}[t]{\x\linewidth}
\centering{\epsfig{figure=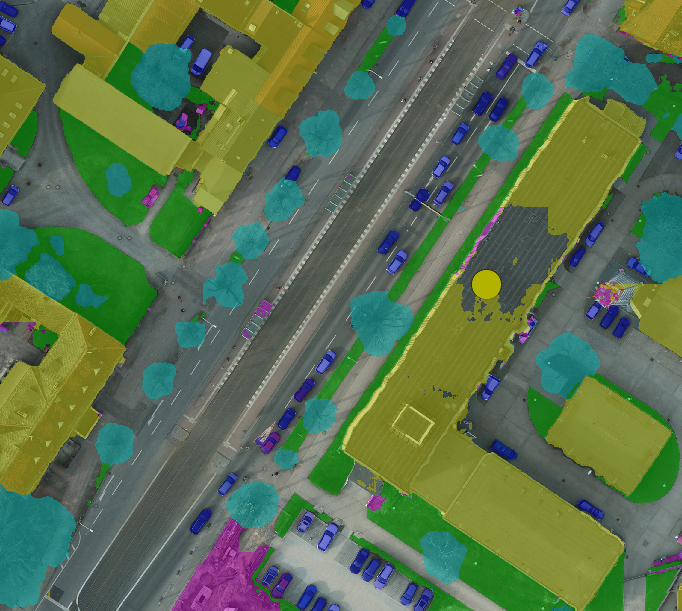,width=\linewidth}}
\end{minipage}\hfill
\begin{minipage}[t]{\x\linewidth}
\centering{\epsfig{figure=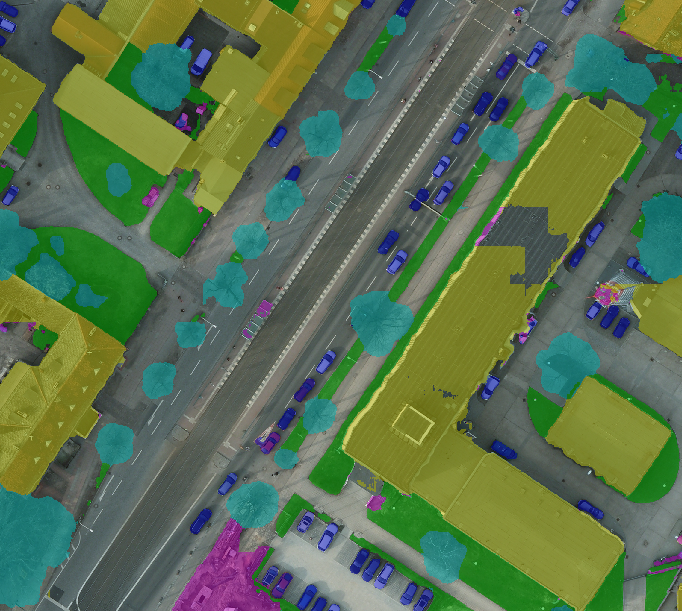,width=\linewidth}}
\end{minipage}\hfill
\begin{minipage}[t]{\x\linewidth}
\centering{\epsfig{figure=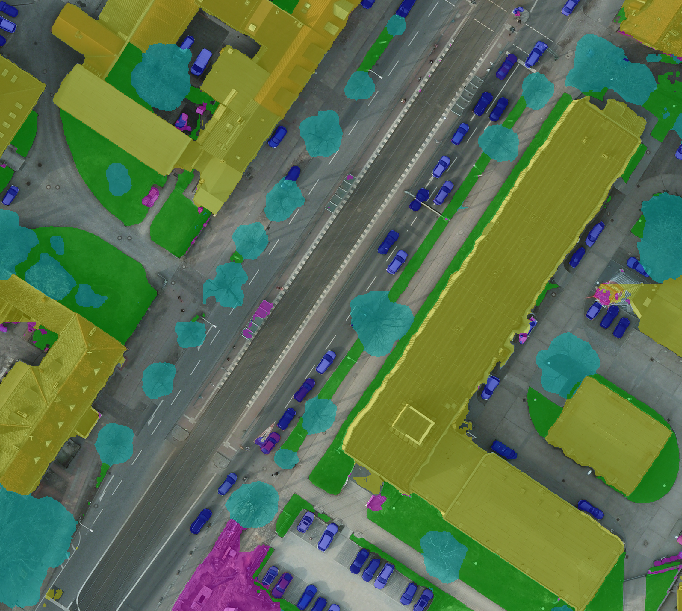,width=\linewidth}}
\end{minipage}

\begin{minipage}[t]{\x\linewidth}
\centering{\epsfig{figure=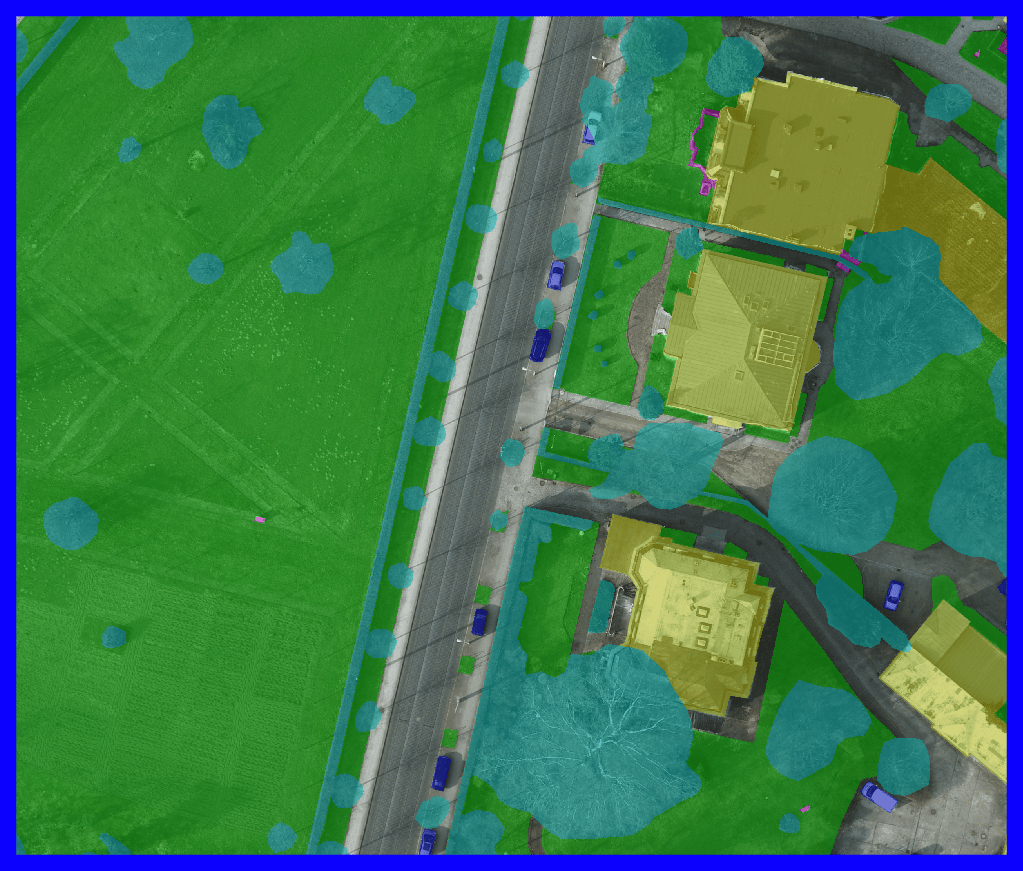,width=\linewidth}}
\end{minipage}\hfill
\begin{minipage}[t]{\x\linewidth}
\centering{\epsfig{figure=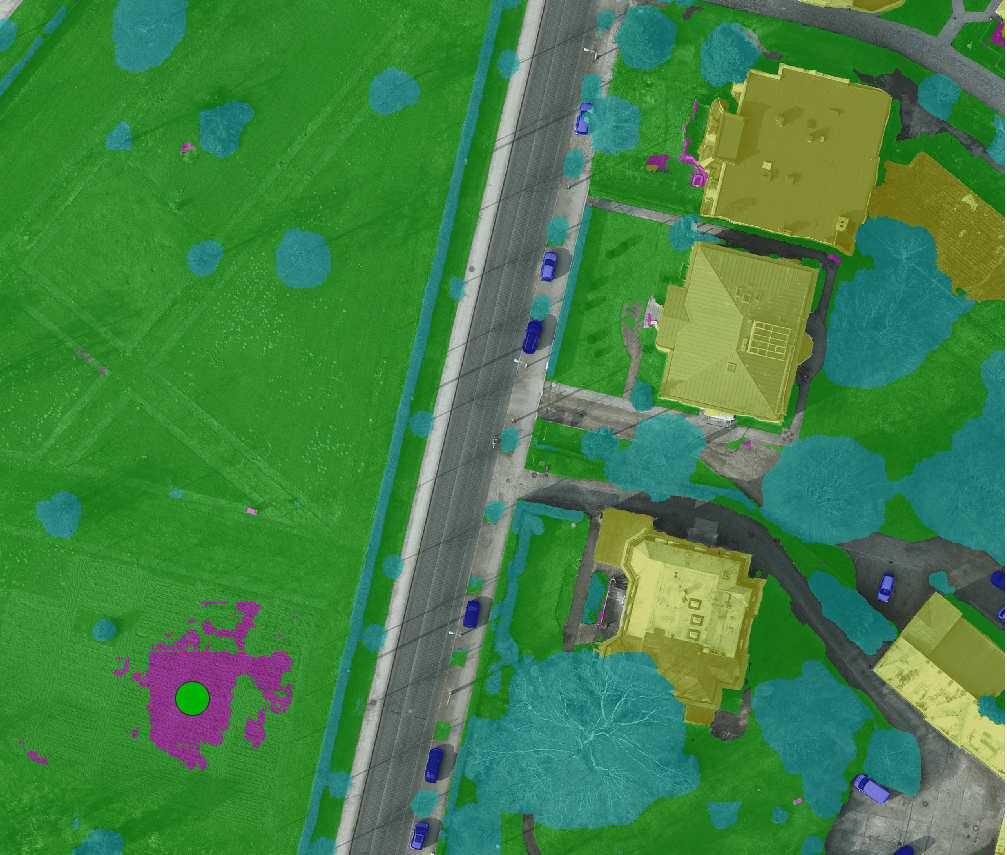,width=\linewidth}}
\end{minipage}\hfill
\begin{minipage}[t]{\x\linewidth}
\centering{\epsfig{figure=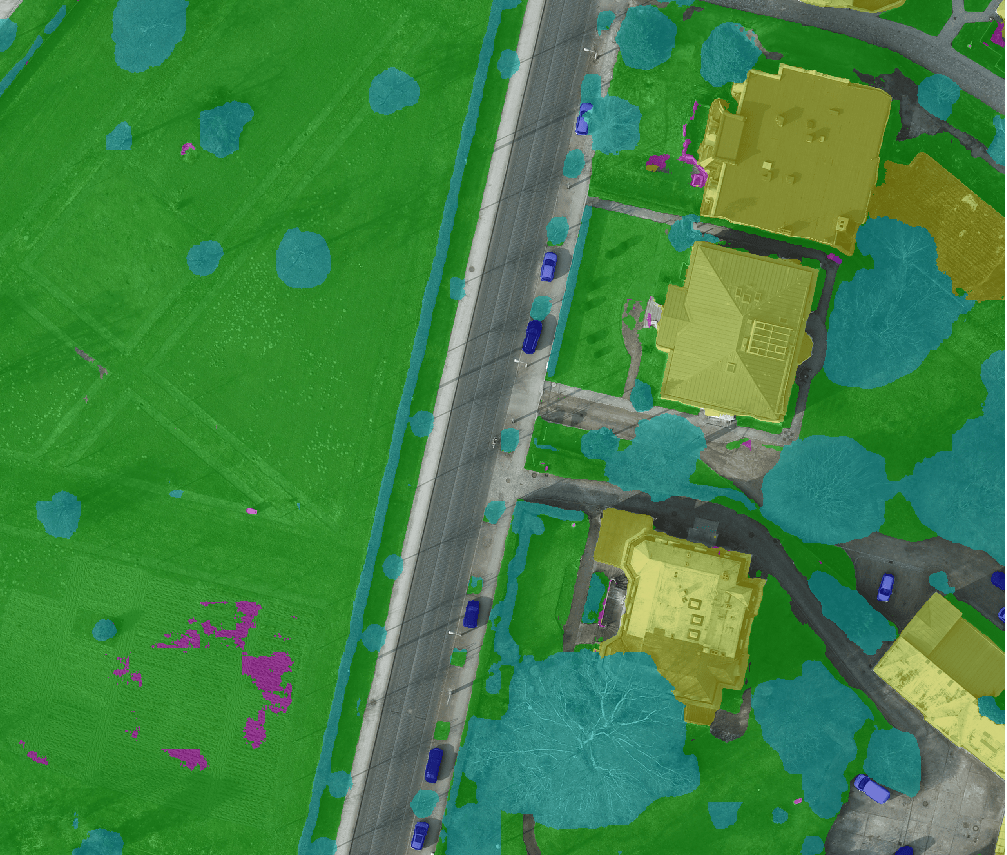,width=\linewidth}}
\end{minipage}\hfill
\begin{minipage}[t]{\x\linewidth}
\centering{\epsfig{figure=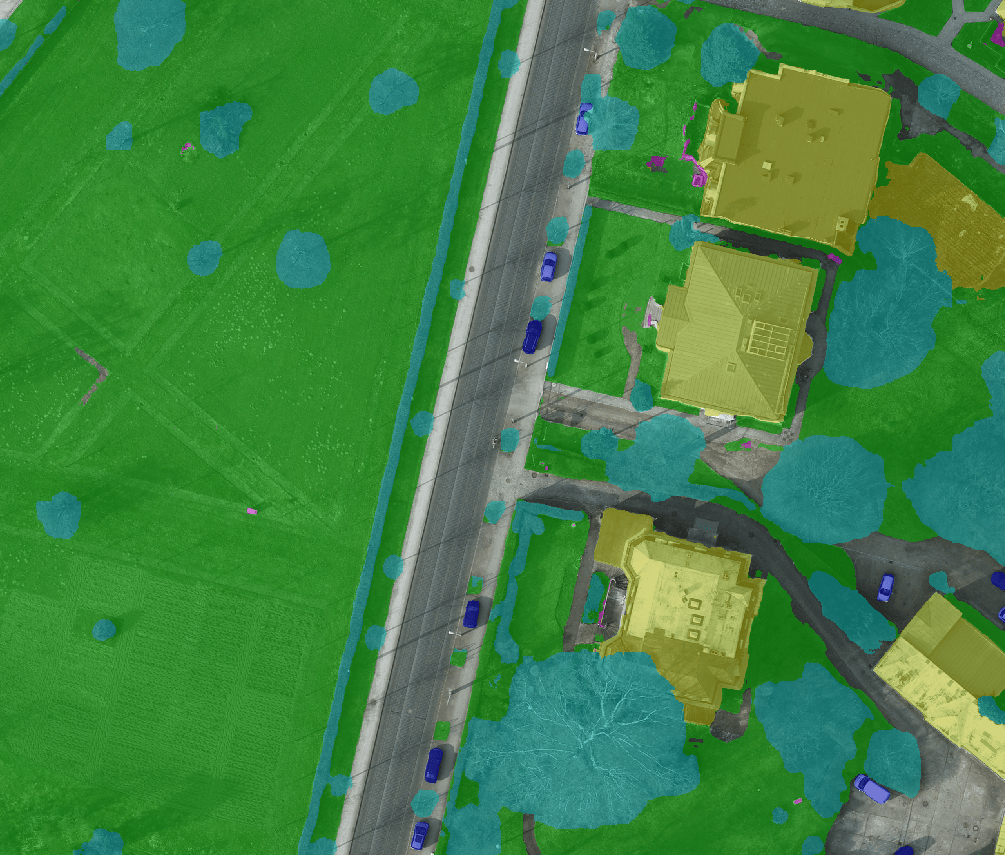,width=\linewidth}}
\end{minipage}

\begin{minipage}[t]{\x\linewidth}
\centering{\epsfig{figure=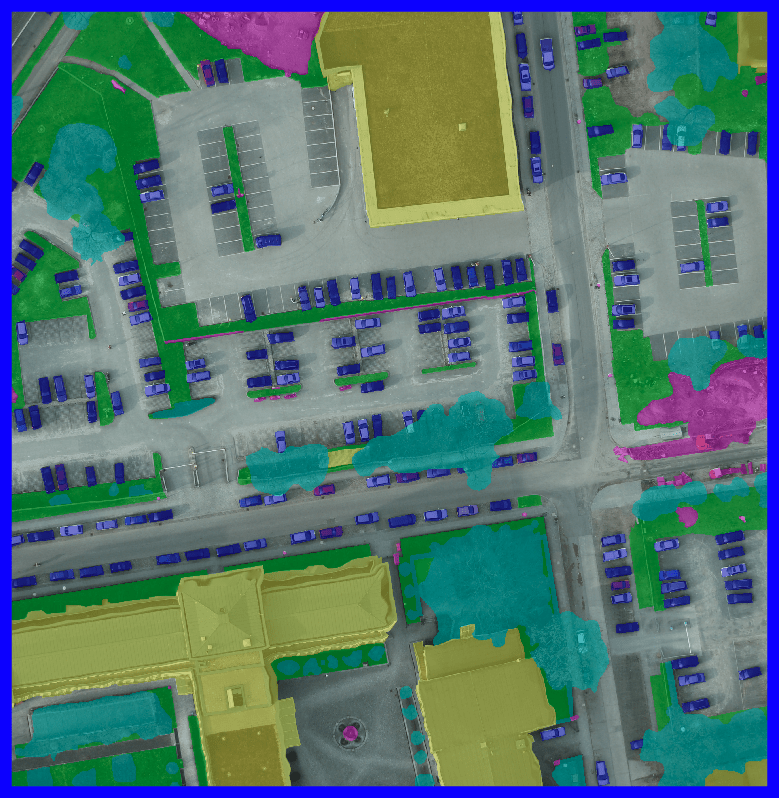,width=\linewidth}}
\end{minipage}\hfill
\begin{minipage}[t]{\x\linewidth}
\centering{\epsfig{figure=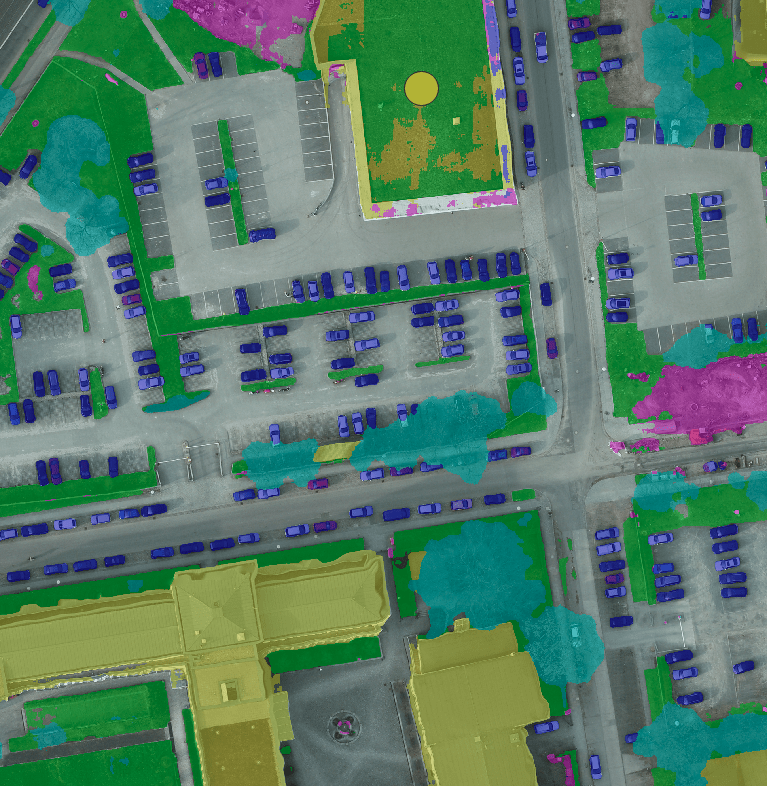,width=\linewidth}}
\end{minipage}\hfill
\begin{minipage}[t]{\x\linewidth}
\centering{\epsfig{figure=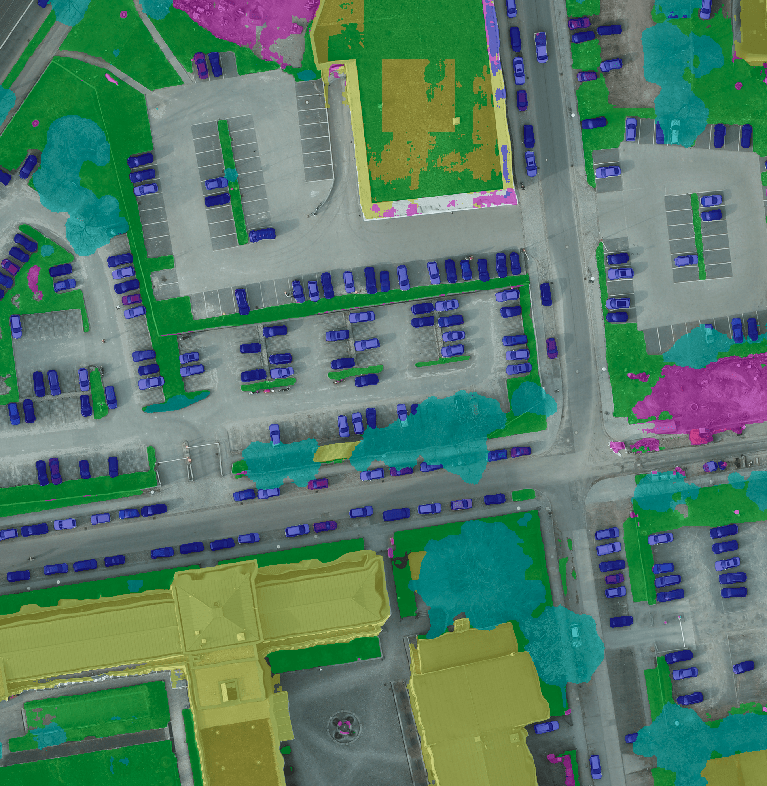,width=\linewidth}}
\end{minipage}\hfill
\begin{minipage}[t]{\x\linewidth}
\centering{\epsfig{figure=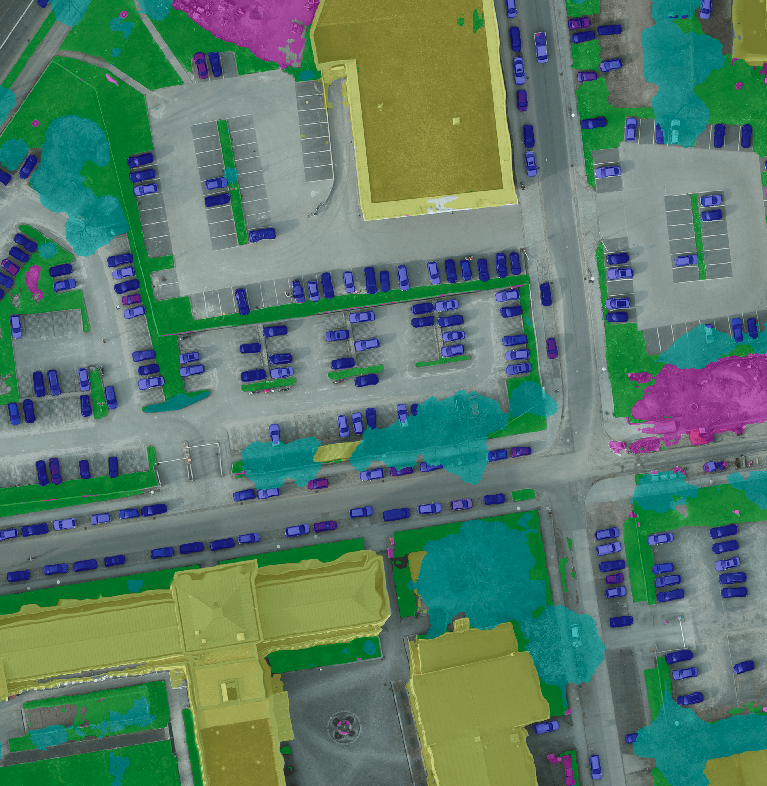,width=\linewidth}}
\end{minipage}

\begin{minipage}[t]{\x\linewidth}
\centering\small{Ground-truth}
\end{minipage}\hfill
\begin{minipage}[t]{\x\linewidth}
\centering\small{Initial prediction with one annotation} 
\end{minipage}\hfill
\begin{minipage}[t]{\x\linewidth}
\centering\small{DISIR}
\end{minipage}\hfill
\begin{minipage}[t]{\x\linewidth}
\centering\small{DISCA}
\end{minipage}
\caption{Results from the ISPRS dataset with single annotations. Best viewed in color.}
\label{fig:toy_pb}
\end{figure}

As shown in Table~\ref{tab:results}, DISCA successfully enhances the initial segmentation maps to reach a higher IoU. Indeed, we observe an average improvement of 2.5\% IoU with ten annotation samples. Besides, it also beats DISIR performances on the three datasets. 

DISCA efficiently allows the user to make corrections at the image scale: on Figure~\ref{fig:toy_pb}, single annotations enable DISCA to provide a corrected segmentation of the scenes while they are not enough for DISIR to deliver a similar result. 

\subsection{Domain adaptation experiment}
\begin{wrapfigure}{R}{0.6\textwidth}
\centering{\epsfig{figure=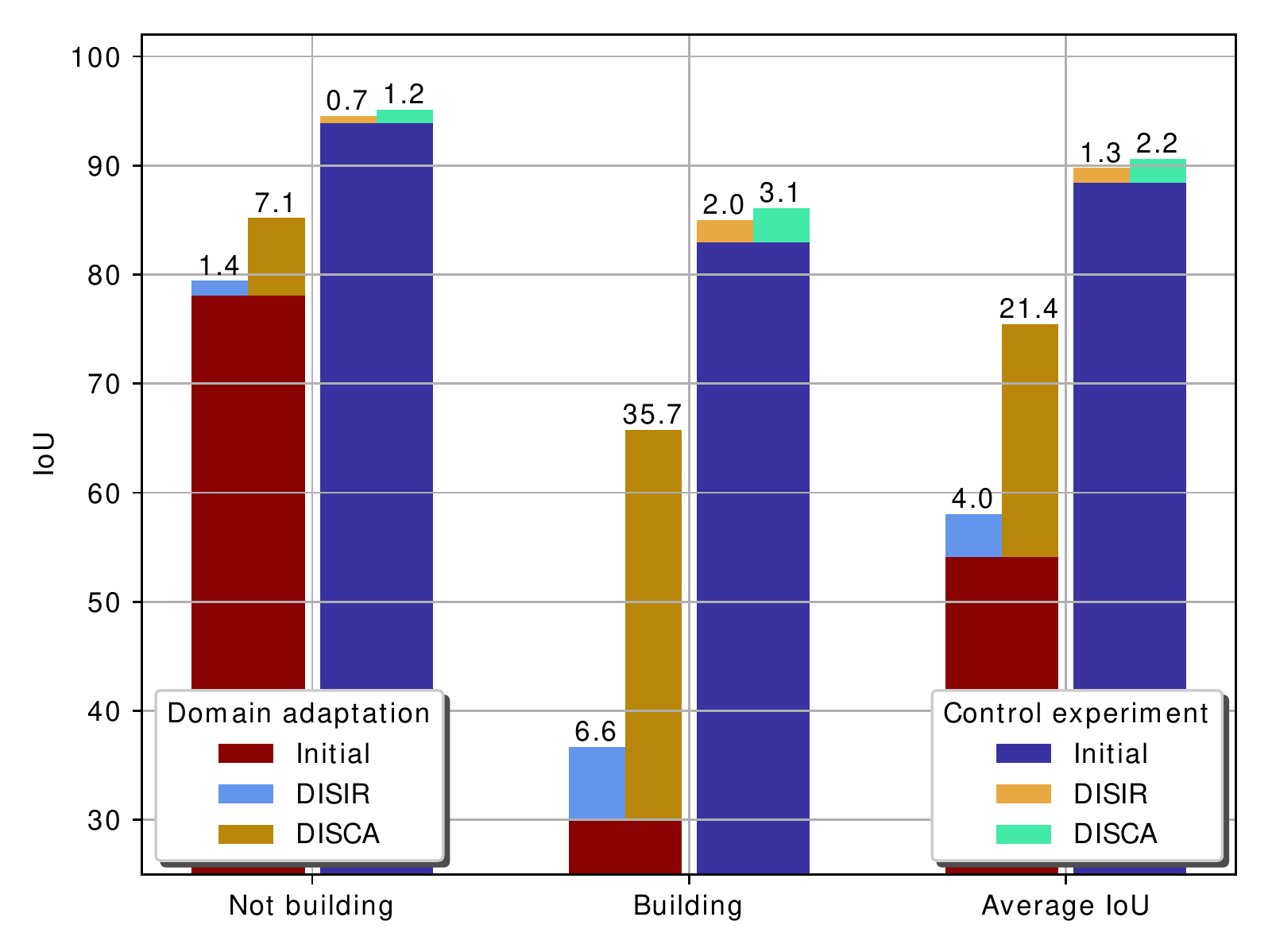,width=0.6\textwidth}}
\caption{Domain adaptation (AIRS$\to$ISPRS)}
\label{fig:domain_adapt}
\end{wrapfigure}
\begin{figure}[h]
\begin{minipage}[t]{\x\linewidth}
\centering{\epsfig{figure=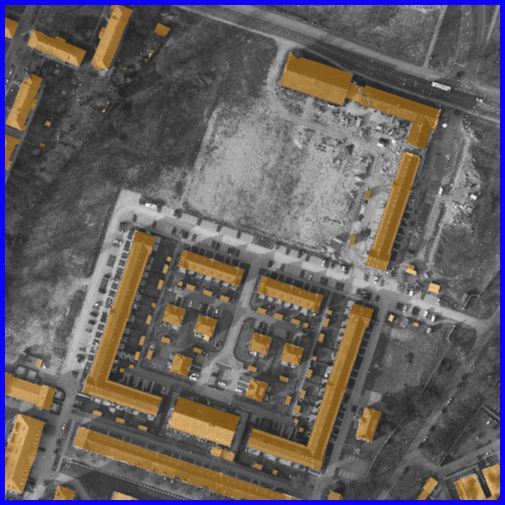,width=\linewidth}}
\end{minipage}\hfill
\begin{minipage}[t]{\x\linewidth}
\centering{\epsfig{figure=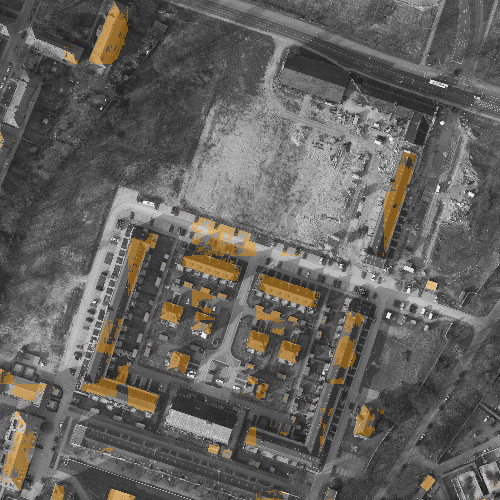,width=\linewidth}}
\end{minipage}\hfill
\begin{minipage}[t]{\x\linewidth}
\centering{\epsfig{figure=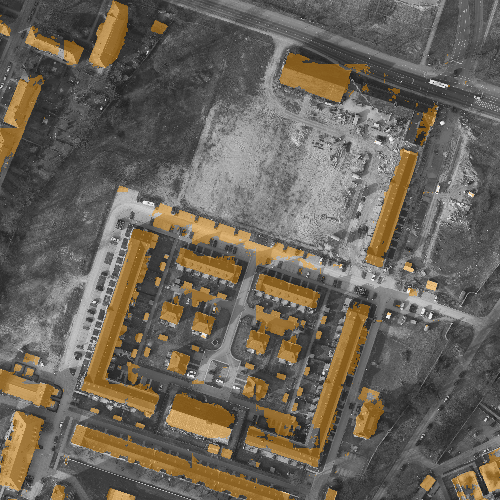,width=\linewidth}}
\end{minipage}\hfill
\begin{minipage}[t]{\x\linewidth}
\centering{\epsfig{figure=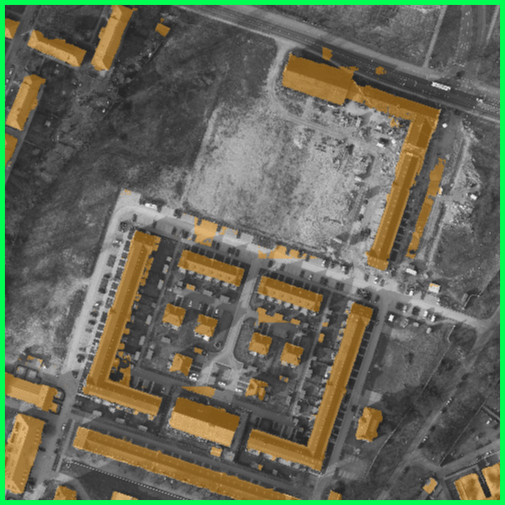,width=\linewidth}}
\end{minipage}

\begin{minipage}[t]{\x\linewidth}
\centering{\epsfig{figure=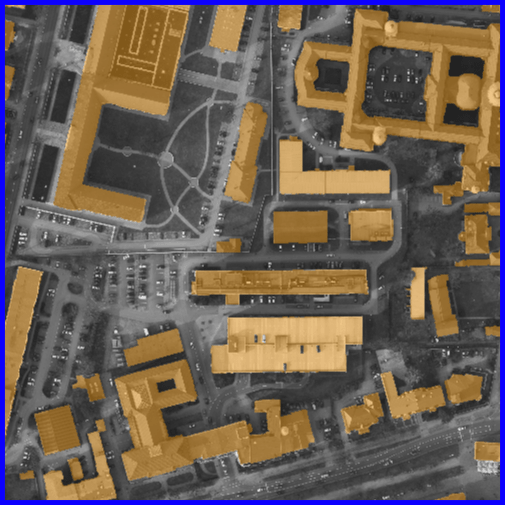,width=\linewidth}}
\end{minipage}\hfill
\begin{minipage}[t]{\x\linewidth}
\centering{\epsfig{figure=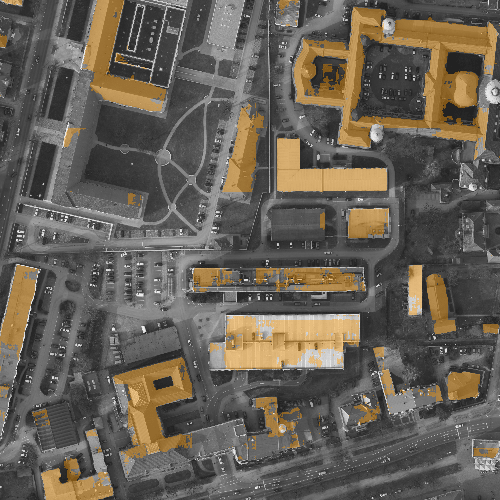,width=\linewidth}}
\end{minipage}\hfill
\begin{minipage}[t]{\x\linewidth}
\centering{\epsfig{figure=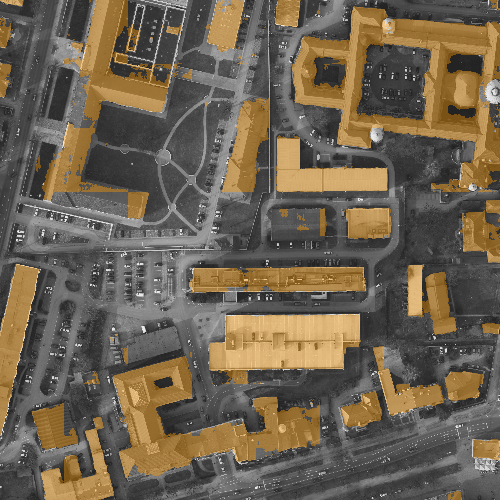,width=\linewidth}}
\end{minipage}\hfill
\begin{minipage}[t]{\x\linewidth}
\centering{\epsfig{figure=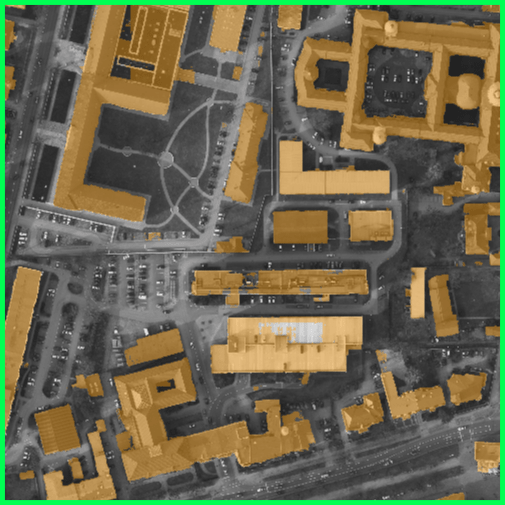,width=\linewidth}}
\end{minipage}

\begin{minipage}[t]{\x\linewidth}
\centering{\epsfig{figure=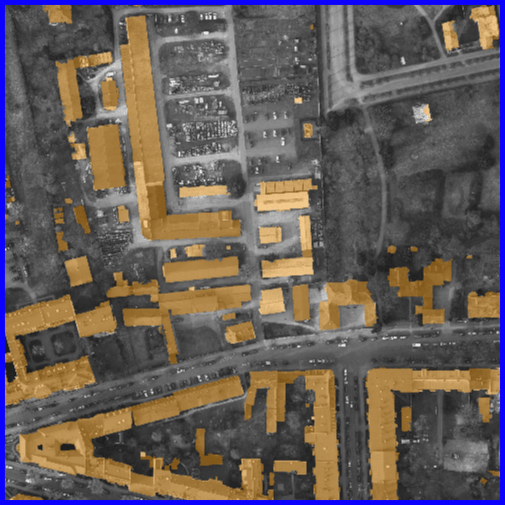,width=\linewidth}}
\end{minipage}\hfill
\begin{minipage}[t]{\x\linewidth}
\centering{\epsfig{figure=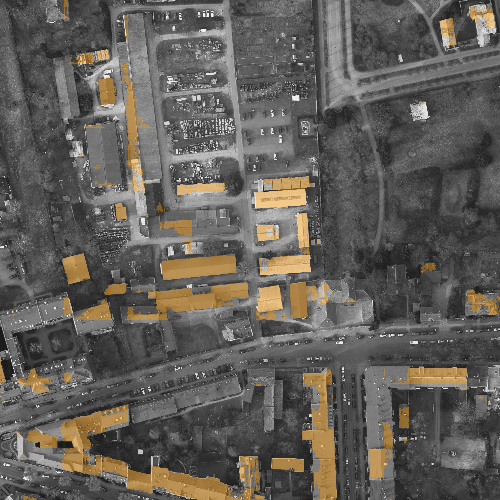,width=\linewidth}}
\end{minipage}\hfill
\begin{minipage}[t]{\x\linewidth}
\centering{\epsfig{figure=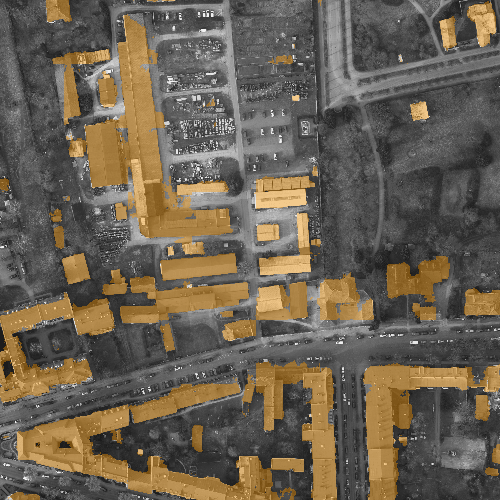,width=\linewidth}}
\end{minipage}\hfill
\begin{minipage}[t]{\x\linewidth}
\centering{\epsfig{figure=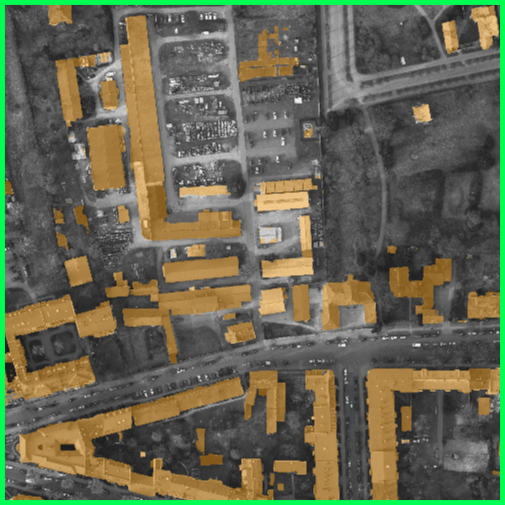,width=\linewidth}}
\end{minipage}

\begin{minipage}[t]{\x\linewidth}
\centering\small{Ground-truth}
\end{minipage}\hfill
\begin{minipage}[t]{\x\linewidth}
\centering\small{Initial prediction} 
\end{minipage}\hfill
\begin{minipage}[t]{\x\linewidth}
\centering\small{Prediction after learning on 10 annotations}
\end{minipage}\hfill
\begin{minipage}[t]{\x\linewidth}
\centering\small{Control prediction}
\end{minipage}
\caption{Building segmentation from the ISPRS validation dataset with a network pretrained on AIRS. The control experiment corresponds to initial predictions made with a network pretrained on ISPRS. Best viewed in color.}
\label{fig:domain_adapt_visual}
\end{figure}

Aiming to push our approach one step further, we evaluated DISCA on ISPRS with a network trained on AIRS to simulate a domain adaptation scenario. The ISPRS labels were simplified to \textit{building} and \textit{not building} and the images were down-sampled using bi-linear interpolation to match the AIRS resolution. For comparison, we also trained a network to detect buildings only on the ISPRS training set and used it as a control experiment. As shown on Figure~\ref{fig:domain_adapt}, even though the performances do not reach the ones from the control experiment, the network within DISCA framework still drastically benefits from the 10 annotations to improve the initially flawed segmentation maps. Indeed, there is a $20\%$ average IoU improvement  with DISCA while it is only around $5\%$ with DISIR. This metric improvement is also visually confirmed on the results displayed on Figure~\ref{fig:domain_adapt_visual} where the segmentation maps are globally well enhanced using DISCA with only ten annotation samples. In particular, the network is then able to adapt to buildings with peculiar roofs or of uncommon size with respect to the AIRS dataset. These outcomes suggest that DISCA is specifically adapted for global enhancements when the neural network is uncertain about the initial prediction.

\section{Conclusion}

We have proposed in this paper an interactive learning strategy to incrementally refine segmentation maps and applied it to EO data. It makes use of user annotations as ground-truth to continuously adapt a neural network to a target image.
Finally, we have showed on three datasets the benefits of our approach and that it can be especially adapted to enhance mitigated initial results when dealing with domain adaptation. 
In the future, we intend to extend our approach to scarce training data and to explore reinforcement policies in order to leverage information provided by the user even better.

%
%
%
\bibliographystyle{splncs04}
\bibliography{bibli}
\end{document}